\documentclass[twoside,leqno,twocolumn]{article}
\usepackage[letterpaper]{geometry}
\usepackage{hyperref} 
\hypersetup{hidelinks}
\usepackage{ltexpprt}

\usepackage{makecell}
\usepackage{multirow}
\usepackage{multicol} 
\usepackage{graphicx}
\usepackage{float}
\usepackage{enumitem}
\usepackage{subfigure}
\usepackage{adjustbox}
\usepackage{makecell}
\usepackage{enumitem}
\usepackage{amsmath}
\usepackage{amssymb}
\usepackage{booktabs}
\usepackage{setspace}

\begin{document}
\setlength{\abovedisplayskip}{5pt}   
\setlength{\belowdisplayskip}{5pt}

\newcommand\relatedversion{}

\title{\Large Treatment-Aware Hyperbolic Representation Learning for Causal
Effect Estimation with Social Networks\relatedversion}
\author{Ziqiang Cui\thanks{City University of Hong Kong, Email:
\{ziqiang.cui,boweihe2-c\}@my.cityu.edu.hk, chenma@cityu.edu.hk} \footnotemark[3]
\and Xing Tang\thanks{FiT, Tencent, China, Email: \{shawntang,sunnyqiao,leocchen,\\xiuqianghe\}@tencent.com}
\and Yang Qiao\footnotemark[2]
\and Bowei He\footnotemark[1]
\and Liang Chen\footnotemark[2]
\and Xiuqiang He\footnotemark[2]
\and Chen Ma\footnotemark[1] \footnotemark[4]
}

\date{}

\maketitle

\renewcommand{\thefootnote}{\fnsymbol{footnote}}
\footnotetext[3]{Work done as an intern in Tencent.} 
\footnotetext[4]{Corresponding author.} 

\fancyfoot[R]{\scriptsize{Copyright \textcopyright\ 2024 by SIAM\\
Unauthorized reproduction of this article is prohibited}}

\begin{abstract} \small\baselineskip=9pt 
Estimating the individual treatment effect (ITE) from observational data is a crucial research topic that holds significant value across multiple domains. How to identify hidden confounders poses a key challenge in ITE estimation. Recent studies have incorporated the structural information of social networks to tackle this challenge, achieving notable advancements. However, these methods utilize graph neural networks to learn the representation of hidden confounders in Euclidean space, disregarding two critical issues: (1) the social networks often exhibit a scale-free structure, while Euclidean embeddings suffer from high distortion when used to embed such graphs, and (2) each ego-centric network within a social network manifests a treatment-related characteristic, implying significant patterns of hidden confounders. To address these issues, we propose a novel method called Treatment-Aware Hyperbolic Representation Learning (TAHyper). Firstly, TAHyper employs the hyperbolic space to encode the social networks, thereby effectively reducing the distortion of confounder representation caused by Euclidean embeddings. Secondly, we design a treatment-aware relationship identification module that enhances the representation of hidden confounders by identifying whether an individual and her neighbors receive the same treatment. Extensive experiments on two benchmark datasets are conducted to demonstrate the superiority of our method. 
The code is available at
\href{https://github.com/ziqiangcui/TAHyper}{https://github.com/ziqiangcui/TAHyper}. 
\end{abstract}
\vspace{-0.6em}
\section{Introduction}
Estimating the individual treatment effect (ITE) from observational data has considerable value across various fields, including healthcare, education, and online marketing. 
The primary challenge of this research area is addressing the confounding issue. Unlike randomized controlled trials where treatments are randomly assigned, observational data often includes confounders. These confounders affect both the treatment assignment and the outcome, leading to a bias in ITE estimation. For example, when estimating the causal effect of an expensive medicine on patients, those with a higher socioeconomic status are more likely to have access to this medication, and their elevated socioeconomic status also affects their health conditions positively. Therefore, socioeconomic status is a confounder in this scenario.

Most of previous studies \cite{johansson2016learning,shalit2017estimating,hassanpour2019learning,alaa2017bayesian,hill2011bayesian,li2017matching,wager2018estimation,yao2018representation,schwab1810perfect} rely on the strong ignorability assumption, which requires all confounders can be measured. However, this assumption is often untenable in real-world scenarios. For example, in the aforementioned scenario, capturing socioeconomic status through individual features such as age and education proves to be extremely challenging \cite{guo2020learning}. Fortunately, the social network information offers an exciting alternative for identifying such hidden confounders, as unobserved confounders like socioeconomic status are often implicitly manifested in social network patterns, such as individuals' community affiliations and social relationships. In this line of research, related studies \cite{guo2020learning,guo2021ignite,ma2021deconfounding,chu2021graph,veitch2019using} leverage the structural information of social networks as a proxy for hidden confounders, achieving notable advancements in the field. Specifically, they employ both the network's adjacency matrix and individual features as input for the model, utilizing Graph Convolutional Networks (GCN) to learn the representation of confounders. The learned representation is subsequently employed for outcome regression, accompanied by a representation balancing term to mitigate confounding bias. 

\begin{figure}
\setlength{\abovecaptionskip}{0cm} 
\setlength{\belowcaptionskip}{-7.5mm} 
	\centering
	\subfigure{
		\begin{minipage}{0.45\linewidth}
			\includegraphics[height=4.0cm, width=0.9\linewidth]{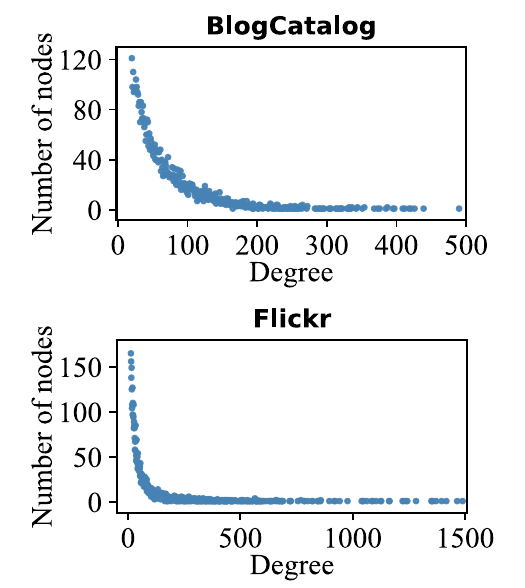}
		\end{minipage}
		\label{intro1}
	}
    	\subfigure{
    		\begin{minipage}{0.45\linewidth}
   		 	\includegraphics[height=4.0cm, width=0.9\linewidth]{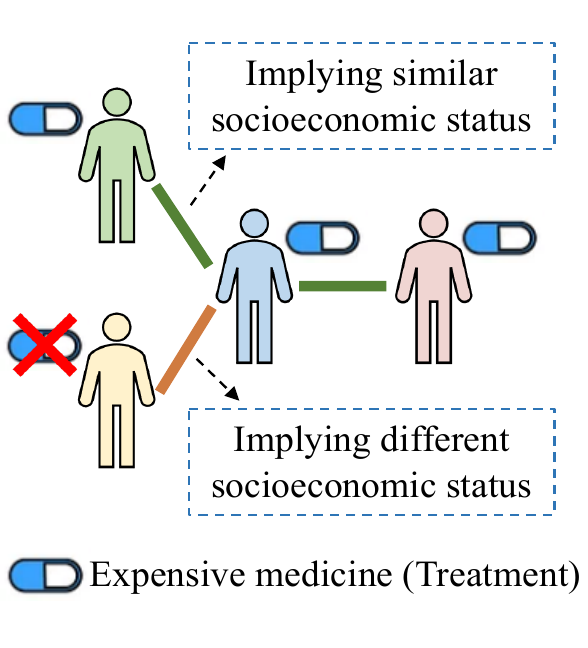}
    		\end{minipage}
		\label{intro2}
    	}
	\caption{Left: Node degree distribution in two benchmark datasets (BlogCatalog and Flickr); Right: An illustrative example of treatment-related relationships.}
	\label{intro}
\end{figure} 
Despite the progress made in utilizing social networks to capture hidden confounders, we argue that existing methods overlook two critical issues:
\vspace{0mm}
\begin{itemize}[leftmargin=*,topsep=0pt,parsep=0pt]
\setlength{\topsep}{0pt}
\setlength{\itemsep}{0pt}
\setlength{\parsep}{0pt}
\setlength{\parskip}{0pt}
\item \textbf{Scale-Free Structure of Social Networks}. 
Real-world social networks often exhibit a scale-free pattern. 
We conduct a statistical analysis on the node degree distribution using two benchmark datasets for ITE estimation with social networks: BlogCatalog and Flickr. 
The results indicate that both datasets demonstrate a power-law distribution (as shown in Fig. \ref{intro}), which meets the criterion for classifying a network as scale-free.
Existing methods utilize graph models to represent the network structure in Euclidean space. However, scale-free networks possess a non-Euclidean latent anatomy \cite{yang2022hyperbolic}. In particular, they have hierarchical tree-like topology \cite{chami2019hyperbolic}, and their volume, defined as the number of nodes within a specific radius of a central node, grows exponentially as the radius increases. In contrast, the volume of Euclidean balls exhibits only a polynomial growth as the radius increases. Moreover, studies have shown that Euclidean embeddings suffer from significant distortion when applied to scale-free networks\cite{chen2013hyperbolicity,chami2019hyperbolic,du2022hakg,sala2018representation}, even with an unbounded number of dimensions \cite{sala2018representation}. 
Existing methods consider the network structure as a proxy for hidden confounders. Thus, accurate modeling of the structural information is crucial for effectively identifying hidden confounders. However, the distorted network representation in Euclidean space can hinder the identification of hidden confounders, thereby undermining the precision of ITE estimation.

\vspace{0mm}
\item \textbf{Treatment-Related Social Relationships}. From an individual view, the ego-centric network \cite{arnaboldi2017online} centered around each individual in the social network manifests a treatment-related characteristic.
Specifically, the neighbors of each individual can be classified into two categories: those receiving the same treatment as the individual and those receiving a different treatment.
Sharing the same treatment usually suggests similar hidden confounders, while having different treatments implies distinct hidden confounders.
To illustrate, Fig. \ref{intro} demonstrates an example in the context of estimating the causal effect of an expensive medicine. We can see the central individual (highlighted in blue) takes the medicine. 
Due to the high cost of the medication, 
individuals with a higher socioeconomic status are more likely to have access to this medication.
Consequently, the neighbors of the central individual who also use this medication (highlighted in green and pink) are highly likely to share a similar socioeconomic status with her, whereas the remaining neighbor (highlighted in yellow) is more likely to have a distinct socioeconomic status.
In other words, these treatment-related relationships imply the pattern of socioeconomic status, which serves as a key hidden confounder in this example. 
Unfortunately, current methods overlook this distinctive characteristic and merely rely on GCN for comprehending the local structure.
To effectively identify hidden confounders, it is crucial to distinguish between these treatment-related relationships that contain significant information about hidden confounders.
\end{itemize} 

To tackle these issues, we propose a novel method named
\textbf{T}reatment-\textbf{A}ware \textbf{Hyper}bolic Representation Learning (\textbf{TAHyper}). TAHyper enhances the ability of identifying hidden confounders with social networks from both the macro and micro perspectives, thus mitigating the confounding bias in ITE estimation.
Our method has two key highlights.
Firstly, we employ the hyperbolic space to model the structure of social networks from a macro view. 
Unlike Euclidean space, hyperbolic space exhibits exponential expansion in proportion to its radius.
This geometry property closely matches the growth rate of scale-free networks, and endows hyperbolic space with ample capacity to preserve the relative positions of nodes and the hierarchical structure, resulting in minimal distortion. 
Accordingly, we design a hyperbolic representation module that employs hyperbolic space to capture hidden confounders implicit in the network structure.
Secondly, recognizing the characteristic of treatment-related relationships, we design a treatment-aware relationship identification module to enhance the representation learning of hidden confounders from an individual view. This module discerns whether an individual's neighbors receive the same treatment as the individual based on their confounder representation. By optimizing the discrimination error, the learned representation can capture the hidden confounder pattern implied in the treatment-related relationships.
Overall, our main contributions can be summarized as follows:
\begin{itemize}[leftmargin=*,topsep=0pt,parsep=0pt,itemsep=0pt]
\item We employ the hyperbolic space to encode the structural information of social networks. 
This approach significantly reduces the representation distortion, thereby improving the ability of using social networks to identify hidden confounders.
\item We design a treatment-aware relationship identification module, which enhances the confounder representation learning by capturing the pattern of hidden confounders implied in treatment-related social relationships.
\item We conduct extensive experiments on two public benchmark
datasets, and the results demonstrate the superiority of our method.
\end{itemize}
\vspace{-0.5em}
\section{Related Work}
Two primary categories of approaches exist for estimating ITE from networked observational data.
\begin{list}{}{\leftmargin=0em \itemindent=0em \topsep=0pt \parsep=0pt \itemsep=0pt}
\item \textbf{Capturing Hidden Confounders.} The assumption of ``no hidden confounding" is often untenable in practice. To address this issue, researchers relax this assumption by utilizing the network structure as a proxy for hidden confounders\cite{veitch2019using,guo2020learning,ma2021deconfounding,chu2021graph,guo2021ignite}. Network Deconfounder \cite{guo2020learning} is the pioneering work that utilizes GCN to capture hidden confounders. Building upon this approach, 
IGNITE \cite{guo2021ignite} proposes a minimax game framework to achieve two objectives: predicting treatment assignments and balancing the representation distributions. Additionally, a dynamic network deconfounder \cite{ma2021deconfounding} is proposed to model the evolution of observational data over time. Recently, GIAL \cite{chu2021graph} is proposed to maximize the mutual information between individual embeddings and the structural information of the entire graph through adversarial learning. However, these studies overlook two crucial factors that influence the confounder identification: the scale-free structure of social networks and treatment-related social relationships.
\item \textbf{Modeling Spillover Effect.} Due to the existence of networks, the potential outcome of individuals is not only affected by their own treatment but also by the treatment of connected individuals. This violates the Stable Unit Treatment Value Assumption (SUTVA). Therefore, this line of studies \cite{ma2022learning, jiang2022estimating,ma2021causal,tran2022heterogeneous,rakesh2018linked} focus on modeling the network interference (i.e., spillover effect).
\end{list}

Our work falls into the first branch. Different from the studies on the spillover effect or network interference, we align with the problem setting of \cite{guo2020learning,guo2021ignite,chu2021graph}, with a focus on leveraging social networks to effectively capture hidden confounders.
\vspace{-0.5em}
\section{Problem Statement}
Firstly, we introduce the networked observational data \cite{guo2020learning}. Let $\mathcal{G({V, E})}$ represent a social network with a vertex set $\mathcal{V}$ and an edge set $\mathcal{E}$. The adjacency matrix $\mathbf{A}\in \mathbb{R}^{n \times n}$ denotes the connections in the network $\mathcal{G}$, where $\mathbf{A}_{ij}=1$ if $(v_i, v_j)\in \mathcal{E}$; otherwise, $\mathbf{A}_{ij}=0$. Here, $n$ represents the total number of nodes. Note that the term ``network" specifically refers to a social network in this paper. 
Generally, in the observational dataset of ITE estimation task, each instance comprises three parts: the features (variables), the treatments, and the outcomes. 
When it comes to networked observational data, the instances are interconnected through the network. Consequently, the networked observational data can be represented as $(\{\mathbf{x}_i, t_i, y_i\}_{i=1}^n, \mathbf{A})$. Here, $\mathbf{x}_i$ denotes the features of the $i$-th individual, $t_i \in \{0,1\}$ represents the observed treatment of the $i$-th individual (considering only binary treatment), where $t_i=1$ ($t_i=0$) indicates that the $i$-th individual is under treatment (control), and $y_i$ denotes the factual (observed) outcome corresponding to $t_i$.

Next, we can estimate the individual treatment effects with the networked observational data. We adopt the potential outcome framework, where the individual treatment effect for the $i$-th individual is defined as the difference between the potential outcomes under treatment and control, denoted as $\text{ITE}_i=y_i^1-y_i^0$. Here, $y_i^1$ ($y_i^0$) represents the potential outcome of the $i$-th individual under treatment (control). Defining ITE, the average treatment effect (ATE) can be formulated as the average of ITE across all instances, given by $\frac{1}{n} \sum_{i=1}^{n}(y_i^1-y_i^0)$. Given $(\{\mathbf{x}_i, t_i, y_i\}_{i=1}^n, \mathbf{A})$, we aim to develop a causal inference model that estimates ITE for each individual.

\begin{figure}[t]
\setlength{\abovecaptionskip}{-1mm} 
\setlength{\belowcaptionskip}{-5mm} 
  \centering
  \includegraphics[width=0.6\columnwidth]{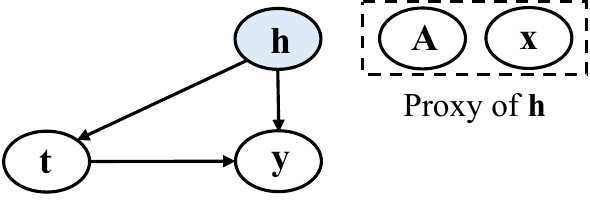}
  \caption{The causal diagram (h is hidden confounders).} 
   \label{causal}
\end{figure}

In the potential outcome framework, many previous studies are based on the strong ignorability assumption:
\vspace{-16pt}
\begin{Definition}
\rm
    \textbf{Strong Ignorability}: Given covariates $x$, treatment $t$ is independent of the potential outcomes, and the probability of each unit to get treated is larger than 0 and less than 1, i.e., $(y^1,y^0) \perp t|x \;and \; 0<Pr(t=1|x)<1$, for all $x$ and $t$.
\end{Definition}
\vspace{-1.5mm}
This assumption is also known as the ``no hidden confounding'' assumption \cite{shalit2017estimating}, which requires all confounders can be measured. 
We relax this assumption by acknowledging the existence of hidden confounders, and use the network structure and individual features as proxies for them, as shown in Fig. \ref{causal}.
In summary, our goal is to utilize the network structure and individual features to learn the representation of hidden confounders and estimate ITE based on the acquired confounder representation.

\begin{figure*}[t]
\setlength{\abovecaptionskip}{-0.5mm} 
\setlength{\belowcaptionskip}{-4mm} 
  \centering
  \includegraphics[width=0.98\textwidth]{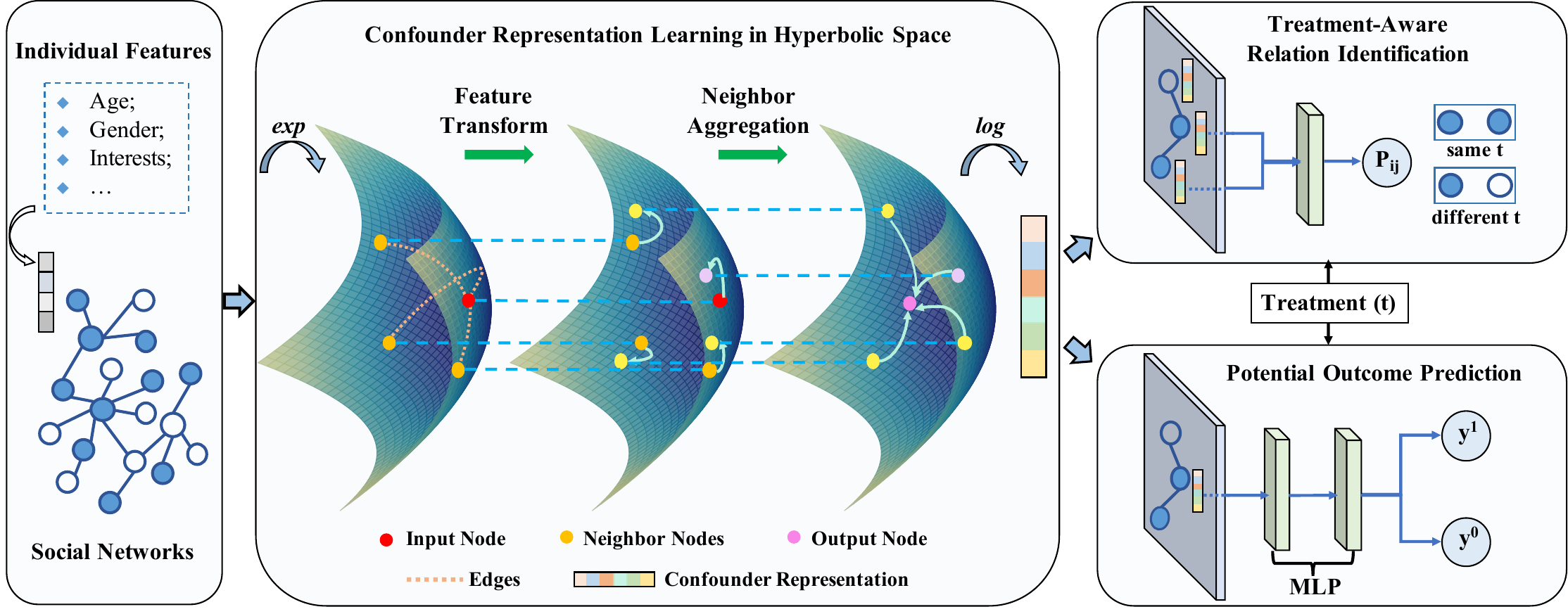}
  \caption{Overview of Our Method (TAHyper). 
  }
   \label{method}
\end{figure*}
\vspace{-0.5em}
\section{Methodology} 
As shown in Fig. \ref{method}, our proposed method consists of three components: hyperbolic confounder representation learning, treatment-aware relationship identification, and potential outcome prediction. 
Firstly, we encode the scale-free structure of social networks as well as individual features in hyperbolic space to learn the representation of hidden confounders.
Next, we employ the treatment-aware relationship identification module to capture the hidden confounder pattern within treatment-related relationships, aiming to enhance the representation learning of hidden confounders.
Finally, we predict potential outcomes by employing two regression networks and a balancing term that addresses the issue of covariate shift, based on the learned confounder representation. 
In the following, we begin by introducing fundamental concepts of hyperbolic space and subsequently delve into the details of each component in our model.
\vspace{-0.8em}
\subsection{Preliminaries\label{sec:preliminaries}}
Hyperbolic geometry is a non-Euclidean geometry with a constant negative curvature. There are five isometric models for hyperbolic geometry, and in this paper, we work with the Poincar\'{e} ball model, one of the extensively used models. 
The Poincar\'{e} ball model is defined as a Riemannian manifold $(\mathbb{B}^d_c, g^\mathbb{B})$, where $\mathbb{B}^d_c=\{{\mathbf{x} \in \mathbb{R}^d: {\Vert \mathbf{x} \Vert}^2<\frac{1}{c}}\}$ is a d-dimensional ball with a radius of $\frac{1}{\sqrt{c}}$ ($c>0$, and the curvature is $-{c}$) \cite{ma2021knowledge}.
The Riemannian metric $g^\mathbb{B}$ is defined as: $ g^\mathbb{B}=\lambda_\mathbf{x}^2g^{E}$,
where $\lambda_\mathbf{x}=\frac{2}{1-c{\Vert \mathbf{x} \Vert}^2}$ is the conformal factor and $g^{E}=\mathbf{I}_d$ is the Euclidean metric.

A tangent space $\mathcal{T}_\mathbf{z}\mathbb{B}$ is a Euclidean space which is defined as the first-order approximation of $\mathbb{B}$ around the point $\mathbf{z}$. To establish a bridge between the tangent space and the hyperbolic space, exponential and logarithmic maps are employed. Specifically, the exponential map maps the tangent space to the hyperbolic space $\mathbb{B}$, while the logarithmic map performs the reverse mapping from the hyperbolic space back to the tangent space.
Given $z \in \mathbb{B}^d_c$, tangent vectors $\mathbf{x} \in \mathcal{T}_\mathbf{z}\mathbb{B}$ and $\mathbf{y} \in \mathcal{T}_\mathbf{z}\mathbb{B}$ ($\mathbf{x}\neq \mathbf{y}$), the exponential map and the logarithmic map are formulated as:
\begin{equation}
\vspace{-0.3em}
\label{eq1}
    \exp _\mathbf{z}^c(\mathbf{x})=\mathbf{x}\oplus_c(\tanh(\sqrt{c} \frac{\lambda_\mathbf{z}^c {\Vert \mathbf{x} \Vert}^2}{2} )\frac{\mathbf{x}}{\sqrt{c{\Vert \mathbf{x} \Vert_2}}} )
    \,,
\end{equation}
\begin{equation}
\vspace{-0.3em}
\label{eq2}
    \log _\mathbf{z}^c(\mathbf{y})= \frac{2}{\sqrt{c\lambda_\mathbf{z}^c}} {\tanh}^{-1}(\sqrt{c}{\Vert -\mathbf{z}\oplus_c \mathbf{y}\Vert_2}) \frac{-\mathbf{z}\oplus_c \mathbf{y}}{{\Vert-\mathbf{z}\oplus_c \mathbf{y}\Vert_2}} 
    \,,
\end{equation}
where $\oplus$ is the M{\"o}bius addition \cite{ganea2018hyperbolic}. Based on these two mappings, a hyperbolic version of graph convolutional networks has been derived \cite{chami2019hyperbolic}, achieving significant success in traditional graph tasks.
\vspace{-0.8em}
\subsection{Hyperbolic Confounder Representation Learning}
Identifying the hidden confounders is the foremost challenge of estimating ITE from observational data. 
Recent studies have demonstrated that the structural information of social networks can serve as a reliable proxy for hidden confounders \cite{guo2020learning,chu2021graph,guo2021ignite,ma2021deconfounding}. Building upon this insight, existing methods employ the graph neural networks to encode the network structure in Euclidean space, aiming to obtain an improved representation of hidden confounders. However, social networks often exhibit a scale-free structure, while Euclidean embeddings suffer from high distortion when applied to scale-free graphs \cite{chami2019hyperbolic,chen2022modeling}. The distorted representation fails to accurately capture the structure of social networks, thereby impeding the learning of hidden confounders that are implied in the network structure.
Fortunately, hyperbolic space provides a compelling alternative. Unlike Euclidean space where the volume grows only polynomially, hyperbolic space exhibits exponential expansion in proportion to its radius. This geometry property closely matches the growth rate of scale-free graphs, leading to minimal distortion. The exponential growth also endows hyperbolic space with ample capacity to preserve the relative positions of nodes and the integrity of the structure. Moreover, hyperbolic space can be regarded as a continuous tree \cite{yang2022hyperbolic}, making it well-suited for modeling the hierarchical tree-like topology of scale-free networks. Inspired by this, we employ the hyperbolic space to learn the hidden confounders implicit in social networks. 

Specifically, we aim to learn the confounder representation for each individual based on the network structure and individual features, which can be formulated as learning a function $f:\mathcal{X} \times \mathcal{A} \rightarrow \mathbb{R}^{d^\prime}$. 
First, we map the original individual features to the Poincar\'{e} ball manifold of hyperbolic space by the exponential map: 
\begin{equation}
\vspace{-0.2em}
\label{eq3}
    \mathbf{x}^\mathbb{B} = \exp_\mathbf{o}^c(\mathbf{x}^E) \,,
\end{equation}
where the superscript $\mathbb{B}$ denotes the Poincar\'{e} ball model defined in Section {\ref{sec:preliminaries}},
$\mathbf{x}^E \in \mathbb{R}^d$ denotes input Euclidean features, which can be considered as a point in $\mathcal{T}_\mathbf{o}\mathbb{B}^d_c$, $\mathbf{o}=\{0,0,...,0\} \in \mathbb{B}^d_c$ is the origin in $\mathbb{B}^d_c$, serving as a reference point for performing tangent space operations, $c$ is the absolute value of the curvature, and exp is the exponential map defined in Equation (\ref{eq1}).

Furthermore, it is necessary to effectively encode both the network structure and individual features simultaneously. Since the adjacency matrix is discrete and sparse, directly concatenating it with dense individual features can not yield satisfactory results. To address this issue, we employ the Hyperbolic Graph Convolutional Networks (HGCN)\cite{yang2022hyperbolic,chami2019hyperbolic}, which leverages both the expressiveness of GCN and hyperbolic geometry to learn inductive node representations for scale-free networks. More precisely, we derive the GCN operations in the Poincar\'{e} model of hyperbolic space, to effectively encode both the structural and semantic information of social networks.
The first step is to perform a linear transformation on the features, which is achieved by learning a weight matrix that maps the input features to a new feature space. In HGCN, transformations of points are performed in the tangent space by leveraging the exp and log maps. Specifically, the hyperbolic linear transform is formulated as:
\begin{equation}
\vspace{-0.2em}
\label{eq4}
    \mathbf{W} \otimes^c \mathbf{x}^\mathbb{B}= {\exp}_\mathbf{o}^c(\mathbf{W}\log_\mathbf{o}^c(\mathbf{x}^\mathbb{B})) \,,
\end{equation}
where $\mathbf{W}$ is a $d^\prime \times d$ weight matrix, $d^\prime$ is the dimension of hidden layers, and the exp and log maps defined in Equation (\ref{eq1}) and Equation (\ref{eq2}) are used to perform Euclidean transformations in the tangent space.
 In addition, the linear feature transformation can be followed by a bias addition through the M{\"o}bius addition \cite{ganea2018hyperbolic}. 
 
 Next, to capture the structural characteristics and relational dependencies within the social networks, we adopt the neighborhood aggregation option defined in Equation (\ref{eq5}), which enables each node to gain a more comprehensive understanding of its local structure.
\begin{equation}
\vspace{-0.2em}
\label{eq5}
    \text{AGG}(\mathbf{x}^\mathbb{B}_i)=\exp_\mathbf{o}^c( \sum_{j\in \mathcal{N}_i }{\alpha_{ij}\log_\mathbf{o}^c(\mathbf{x}_i^\mathbb{B})}) \,,
\end{equation}
where $\text{AGG}(\mathbf{x}^\mathbb{B}_i)$ is the neighborhood aggregation representation of individual $i$, 
${\mathcal{N}_i}$ denotes the neighbor set of individual $i$, 
and $\alpha_{ij}$ denotes the weight of neighbor $j$ to individual $i$, which is calculated as $ \alpha_{ij}=\frac{1}{\sqrt{\hat d_i\hat d_j}}$, where $\hat d_i=\tilde d_i+1$ and $\tilde d_i$ denotes the node degree.
To learn more complex patterns, a non-linear activation function can be utilized as:
\begin{equation}
\vspace{-0.2em}
\label{eq6}
    \sigma^{\otimes ^{{c}}}(\mathbf{x}^\mathbb{B})=
    \exp_\mathbf{o}^{c}
    (\sigma \log_\mathbf{o}^{c}(\mathbf{x}^\mathbb{B})) \,,
\end{equation}
where $\sigma$ denotes the non-linear activation function and ReLU \cite{glorot2011deep} is adopted in this paper.

To summarize, we learn the confounder representation by stacking multiple HGCN layers, and each layer can be formulated as:
\begin{equation}
\vspace{-0.2em}
\label{eq7}
    \mathbf{x}^{\ell,\mathbb{B}}_i 
    = \sigma^{\otimes ^{{c}}} (\text{AGG}^{c}(\mathbf{W}^{\ell} \otimes ^{c} \mathbf{x}^{\ell-1,\mathbb{B}}_i \oplus ^{c}\mathbf{b}^\ell)) \,,
\end{equation}
where the superscript $\ell$ denotes the $\ell$-th layer, and 
the input of the first layer ($\ell=1$) is defined in Equation (\ref{eq3}). 
In addition, we denote the output of the final layer $\mathbf{x}^{\ell_{m},\mathbb{B}}_i$ as $\mathbf{h}^\mathbb{B}_i$ for simplification, where $\ell_{m}$ is the total number of layers.
\vspace{-0.8em}
\subsection{Treatment-Aware Relationship Identification}
From an individual view, the ego-centric network centered around each individual exhibits a treatment-related characteristic.
Specifically, the neighbors of each individual can be divided into two categories: those receiving the same treatment as the individual and those receiving a different treatment. These relationships encompass crucial information about hidden confounders that can be leveraged to enhance the confounder learning.
In the aforementioned example, when an individual takes the expensive medicine, her neighbors who also take this medicine are more likely to share similar socioeconomic status with her, whereas other neighbors in the control group are more likely to have a distinct socioeconomic status. Therefore, distinguishing between these two types of social relationships can indirectly reveal the underlying confounder pattern, thereby enhancing the representation learning of hidden confounders.
To accomplish this objective, we design a treatment-aware relationship identification module, which can be formulated as learning a function $g: \mathbb{R}^{2d^\prime} \rightarrow \mathbb{R}$.

Specifically, we map the output of the hyperbolic representation module to the tangent space with the logarithmic map, as shown in Equation (\ref{eq8}), and perform the Euclidean relationship classification. Note that another alternative approach is to directly identify the relationships within the Poincar\'{e} ball manifold, and recent studies \cite{chami2019hyperbolic} have shown these two methods exhibit similar performance.
\begin{equation}
\vspace{-0.2em}
\label{eq8}
    \mathbf{h}^E = \log_{\mathbf{o}}^c (\mathbf{h}^\mathbb{B}) \,.
\end{equation}

Assuming that individual $i$ has a neighbor $j \,(j \in \mathcal{N}_i)$, we concatenate their Euclidean representations and then feed the resulting concatenation into a Multi-Layer Perceptron (MLP) layer to generate a probability. This probability indicates the likelihood that individuals $i$ and $j$ receive the same treatment, i.e., $t_i=t_j$. The computation of the probability $\hat{p}_{ij}$ is formulated as:
\begin{equation}
\vspace{-0.2em}
\label{eq9}
    \hat{p}_{ij} = \text{sigmoid}(\mathbf{W}^I(\mathbf{h^E_i} \,||\, \mathbf{h^E_j}) + {b}^I) \,,
\end{equation}
where $\mathbf{h^E_i}$ and $\mathbf{h^E_j}$ denote the Euclidean representations of individual $i$ and its neighbor $j$, respectively, and $(\mathbf{h^E_i} \,||\, \mathbf{h^E_j})$ represents the concatenation of them, $\mathbf{W}^I \in \mathbb R^{1\times 2d^\prime} $ is the learnable weights and ${b}^I$ is the bias item. 

Then, let $p_{ij}$ denote the true relationship type of a pair of individuals, which is defined as:
\begin{equation}
\label{eq10}
 \vspace{-0.2em}
    \begin{aligned} 
    p_{ij} &= \begin{cases} 1, &\text{if } t_i=t_j \\ 
    0, &\text{if } t_i \neq t_j \end{cases} \\
    \end{aligned} \,.
\end{equation} 
We aim to identify these two different types of treatment-related relationships based on the confounder representation. The cross entropy loss is adopted as:
\begin{equation}
\vspace{-0.4em}
\label{eq11}
    \mathcal{L}_t = - \frac{1}{|\mathcal{E}|} \sum_{i=1}^{n}  \sum_{j \in \mathcal{N}_i} 
    p_{ij} \log(\hat{p}_{ij}) + (1-p_{ij}) \log(1-\hat{p}_{ij}) \,,
\end{equation}
where $|\mathcal{E}|$ is the edge number and $n$ is the node number.

Note that this task differs from predicting the treatment assignment for each individual. Instead, our focus is on determining whether the treatments of two connected individuals are identical. Therefore, this task does not conflict with the representation balancing task in Section \ref{sec4.4}.
Furthermore, we employ a sampling strategy to reduce memory usage. Let $\mathcal{N}_i^1$ and $\mathcal{N}_i^0$ represent the neighbors of individual $i$ under treatment $t=1$ and treatment $t=0$, respectively. We randomly select $n_i^s$ neighbors from both $\mathcal{N}_i^1$ and $\mathcal{N}_i^0$, where
$n_i^s = \lceil \frac{1}{5}{\mathop{min}(|\mathcal{N}_i^1|, |\mathcal{N}_i^0|)} \rceil$. The sampling is performed before the training phase, and we can switch to different samples for each individual in different epochs to enhance the training process.

\vspace{-0.8em}
\subsection{Potential Outcome Prediction} \label{sec4.4}
\begin{list}{}{\leftmargin=0em \itemindent=0em \topsep=0pt \parsep=0pt \itemsep=2pt}
\item \textbf{Outcome Regression.}
We utilize deep neural networks to map the confounder representation as well as the treatments to potential outcomes. This mapping can be formulated as a function $\varphi: \mathbb{R}^{d^\prime} \times \{0,1\} \rightarrow \mathbb{R}$. 
Following the approach of TARNet \cite{shalit2017estimating}, we design two regression networks $\varphi^1$ and $\varphi^0$, one for each treatment arm. 
For each individual $i$, the potential outcomes are computed as:
\begin{equation}
\vspace{-0.3em}
\label{eq12}
    \begin{aligned} 
    \varphi({\mathbf{h}_i^E, t}) 
    &= \begin{cases} 
    \varphi^1({\mathbf{h}_i^E}), &\text{if } t=1 \\ 
    \varphi^0({\mathbf{h}_i^E}), &\text{if } t=0
    \end{cases} \\
    \end{aligned} \,,
\end{equation} 
where $\mathbf{h}_i^E$ denots the Euclidean confounder representation. The regression networks are composed of multiple hidden layers and non-linear activation functions, followed by a regression output layer. Note that the parameters are not shared between $\varphi^1$ and $\varphi^0$.
Thus far, we have demonstrated the entire process of inferring the potential outcome $y_i^{t}$ for individual $i$ under treatment $t$, which can be formulated as:
\begin{equation}
\label{eq13}
    \hat{y}_i^{t} =  \varphi({\log_{\mathbf{o}}^c(f(\mathbf{x}_i^E, \mathbf{A})), t}) \,.
\end{equation}

In the training phase, our goal is to minimize the discrepancy between the factual outcomes and the inferred factual outcomes. The loss function of mean squared error is adopted as follows,
\begin{equation}
\label{eq14}
    \mathcal{L}_y = \frac{1}{n} \sum _{i=1}^{n} {(\hat{y}_i^{t_i} - y_i)^2} \,,
    \vspace{-0.5em}
\end{equation}
where $\hat{y}_i^{t_i}$ and ${y}_i$ are the inferred factual outcomes and the observational factual outcomes of individual $i$, respectively. 
Note that we only use the factual outcomes of each individual as the training labels, as we can only observe the factual outcomes with respect to the factual treatment in the observational data. 
\vspace{-0.2em}
\item \textbf{Representation Balancing}. Estimating ITE requires the model to predict the potential outcomes of the same unit under different treatments. However, the training data only includes the factual outcomes for each instance, while the counterfactual outcomes remain unobservable. 
In other words, the regression model $\varphi$ is trained on $(\mathbf{x}_i^E, \mathbf{A}, t_i)$, but the task involves inferring the potential outcome based on $(\mathbf{x}_i^E, \mathbf{A}, 1-t_i)$, which leads to the covariate shift problem \cite{johansson2016learning}. 
Following previous work \cite{shalit2017estimating}, we solve this issue by introducing the integral probability metric (IPM) to measure the distance between the treated and control distributions, and aim to minimize the IPM during the training phase. The IPM simplifies to the Wasserstein distance $\mathcal{D}(k,q)$ when $\mathcal{Z}$ is the 1-Lipschitz functional space, defined as:
\begin{equation}
\vspace{-0.2em}
\label{eq15}
\mathcal{D}(k,q)= \mathop{inf}\limits_{M \in \mathcal{M}_{k,q}}  \int_{
\mathbf{h}^E \in \{ \mathbf{h}^E_i \}_{i:t_i=1}
} \Vert M(\mathbf{h}^E)-\mathbf{h}^E \Vert k(\mathbf{h}^E)d\mathbf{h}^E \,,
\end{equation}
where $k(\mathbf{h}^E)=Pr(\mathbf{h}^E|t_i=1)$ and $q(\mathbf{h}^E)=Pr(\mathbf{h}^E|t_i=0)$ denote the distributions of the confounder representation in two groups, respectively.  
$\mathcal{M}=\{
M|M:\mathbb{R}^{d^\prime} \rightarrow \mathbb{R}^{d^\prime}   
s.t. \, q(M(\mathbf{h}^E)) = k(\mathbf{h}^E)
\}$
denotes the set of functions that are capable of transforming the representation distribution of the treated group to that of the controlled group. In addition, we utilize the algorithm proposed by \cite{cuturi2014fast} to efficiently approximate this distance.
\vspace{-0.8em}
\subsection{Objective Function}
By integrating the previously mentioned modules, we formulate the ultimate objective function of our model as a weighted sum of the outcome regression loss $\mathcal{L}_y$, the relationship identification loss $\mathcal{L}_t$, the distribution distance $\mathcal{D}(k,q)$, and a regularization term. The formulation is as follows,
\begin{equation}
\label{eq16}
    \mathcal{L} = \mathcal{L}_y + \alpha \mathcal{L}_t + \beta \mathcal{D}(k,q) + \lambda \Vert \theta \Vert_2^2 \,,
\end{equation}
where $\Vert \theta \Vert_2^2$ denotes the squared $l2$ norm regularization term on parameters $\theta$, and
$ \alpha, \beta, \lambda$ are hyperparameters used to control the trade-off among the four terms.
\end{list}

\section{Experiments}
\subsection{Dataset Description}
Following previous studies \cite{guo2020learning,chu2021graph,guo2021ignite}, we conduct our experiments on two benchmark datasets, including BlogCatalog and Flickr, which are both semi-synthetic datasets based on real-world social media graph data. 
\begin{list}{}{\leftmargin=0em \itemindent=0em \topsep=0pt \parsep=0pt}
\item \textbf{BlogCatalog.} 
BlogCatalog is an online social platform where users can publish their blogs. In this dataset, each instance corresponds to a blogger, and each edge represents a social connection between two bloggers. Each blogger's features are represented as bag-of-words, which consists of keywords extracted from their descriptions. We adopt the procedures and settings described in \cite{guo2020learning, chu2021graph} to synthesize the treatments and outcomes. The outcomes represent readers' opinions on each blogger, while the treatments indicate whether a blogger's content is predominantly viewed on mobile devices or desktops. Since the treatment assignments are binary, the bloggers are either assigned to the treatment group or the control group. The bloggers whose blog content is read more on mobile devices are considered as receiving the treatment, whereas those whose content is read more on desktops are considered as being in the control group. The objective is to estimate the individual treatment effect of having more views on mobile devices on readers' opinions. We generate three dataset settings with $k=1,2,4$, representing the extent of confounding bias arising from a blogger's neighbors' topics. A larger value of $k$ indicates a greater bias resulting from the influence of neighbors \cite{guo2020learning}. Further details about the data synthesis can be found in \cite{guo2020learning}.

\textbf{Flickr.} Flickr is a popular online photo management and sharing platform. In this dataset, each instance corresponds to a user, and the edges represent social connections between pairs of users. Each user's features consist of a list of interest tags. The detailed settings and assumptions of this dataset align with those of BlogCatalog, as described in  \cite{guo2020learning, chu2021graph}.
\end{list}

We present an overview of these two datasets in Table \ref{dataset}. Note that we perform the aforementioned simulation 10 times for each setting of $k$ on both datasets.
\begin{table}[t] 
\renewcommand\arraystretch{1.0}
	\centering
	\caption{Dataset Description.}  
 \setlength\tabcolsep{4.2pt}
 \scalebox{0.9}{
\begin{tabular}{ccccc}
\toprule
Datasets & Nodes  & Edges  & Features & Treatments  \\
\midrule
BlogCatalog &  5,196  & 173,468 & 8,189 & 2   \\
Flickr & 7,575  & 239,738 &12,047 &2 \\
\bottomrule
\end{tabular}}
\vspace{-1.7em}
\label{dataset}
\end{table}

\begin{table*}[t]  
\renewcommand\arraystretch{1.0}
	\centering
	\caption{Performance comparison of different methods on BlogCatalog and Flickr datasets.}  
	\label{tab:methodcompare}
	\setlength\tabcolsep{3pt}
 \scalebox{0.9}{
	\begin{tabular}{ccccccccccccc}
		\toprule
		\multirow{3}*{Model} 
            & \multicolumn{6}{c}{BlogCatalog} &\multicolumn{6}{c}{Flickr} \\ 
		\cmidrule(lr){2-7}  \cmidrule(lr){8-13} 
		  & \multicolumn{2}{c}{k=1} &\multicolumn{2}{c}{k=2} &\multicolumn{2}{c}{k=4}& \multicolumn{2}{c}{k=1} &\multicolumn{2}{c}{k=2} &\multicolumn{2}{c}{k=4}\\ 
    \cmidrule(lr){2-3}  \cmidrule(lr){4-5}  \cmidrule(lr){6-7}  \cmidrule(lr){8-9}  \cmidrule(lr){10-11}  \cmidrule(lr){12-13} 
		& $\sqrt{\epsilon _{PEHE}}$ & $\epsilon _{ATE}$ & $\sqrt{\epsilon _{PEHE}}$ & $\epsilon _{ATE}$ & $\sqrt{\epsilon _{PEHE}}$ & $\epsilon _{ATE}$& $\sqrt{\epsilon _{PEHE}}$ & $\epsilon _{ATE}$ & $\sqrt{\epsilon _{PEHE}}$ & $\epsilon _{ATE}$ & $\sqrt{\epsilon _{PEHE}}$ & $\epsilon _{ATE}$\\
		\midrule 
		BART & 5.830 & 2.306 & 11.072 & 5.963 & 26.413 & 10.618 & 10.032 & 6.508 & 13.905 & 9.306 & 27.485 & 13.209 \\
TARNeT & 10.836 & 5.162 & 24.696 & 12.170 & 46.059 & 22.242 & 15.578 & 3.800 & 32.589 & 8.503 & 64.847 & 12.167 \\
 CFR-MMD & 10.068 & 5.113 & 24.048 & 12.124 & 46.679 & 21.923 & 15.184 & 4.642 & 31.289 & 8.259 & 60.758 & 13.587 \\
 CFR-Wass & 9.923 & 5.021 & 23.658 & 11.795 & 45.294 & 21.045 & 15.024 & 4.619 & 31.624 & 8.012 & 59.625 & 12.964 \\
CEVAE & 8.777 & 1.938 & 20.910 & 4.819 & 43.352 & 12.594 & 14.747 & 2.895 & 31.051 & 5.165 & 59.716 & 7.570 \\
 ND & 4.575 & 1.099 & 9.416 & 2.600 & 21.053 & 5.583 & 6.024 & 1.179 & 9.895 & 3.109 & 20.341 & 4.693 \\ 
GIAL & 4.305 & 0.836 & 8.914 & 1.720 & 19.031 & 5.761 & 5.791 & 1.083 & 9.269 & 2.142 & 18.259 & 5.099 \\
NetEst & 4.903 & 0.833 & 10.219 & 1.359 & 24.582 & 5.581 & 10.735 & 0.677 & 17.658 & 1.623 & 36.622 & 5.167 \\
\midrule
\textbf{TAHyper} & \textbf{4.009} & \textbf{0.165} & \textbf{8.030} & \textbf{0.775} & \textbf{18.019} & \textbf{3.494} & \textbf{5.318} & \textbf{0.624} & \textbf{8.638} & \textbf{1.097} & \textbf{17.276} & \textbf{4.123} \\
		\bottomrule
	\end{tabular} }
 \vspace{-1.5em}
\label{comparison}
\end{table*}

\begin{table}[t]  
\renewcommand\arraystretch{0.9}
	\centering
	\caption{Ablation study (BC stands for BlogCatalog).}  
	\label{tab:methodcompare}
	\setlength\tabcolsep{2.8pt}
  \scalebox{0.88}{
	\begin{tabular}{lcccccc}
		\toprule
		\multirow{2}*{} 
  & \multicolumn{2}{c}{k=1} &\multicolumn{2}{c}{k=2} &\multicolumn{2}{c}{k=4} \\ 
		\cmidrule(lr){2-3}  \cmidrule(lr){4-5}  \cmidrule(lr){6-7}
  & $\sqrt{\epsilon _{PEHE}}$ & $\epsilon _{ATE}$ & $\sqrt{\epsilon _{PEHE}}$ & $\epsilon _{ATE}$ & $\sqrt{\epsilon _{PEHE}}$ & $\epsilon _{ATE}$
		  \\ 
\midrule 
\multicolumn{7}{l}{\textbf{BC}} \\
{TAHyper} & {4.009} & {0.165} & {8.030} & {0.775} & {18.019} & {3.494} \\
\makecell[l]{w/o HB} & {4.355} & {0.923} & {8.439} & {2.440} & {19.203} & {5.580} \\
\makecell[l]{w/o TA} & {4.284} & {0.587} & {8.557} & {1.076} & {18.768} & {4.845} \\
\midrule
\multicolumn{7}{l}{\textbf{Flickr}} \\
{TAHyper} & {5.318} & {0.624} & {8.638} & {1.097} & {17.276} & {4.123} \\
\makecell[l]{w/o HB} & {5.551} & {0.564} & {9.053} & {1.349} & {17.962} & {4.503} \\
\makecell[l]{ w/o TA} & {5.513} & {0.812} & {9.172} & {1.137} & {17.824} & {4.285} \\
		\bottomrule
	\end{tabular} }
 \label{ablation}
 \vspace{-1.2em}
\end{table}

\vspace{-0.5em}
\subsection{Baseline Methods}
We compare our method with the following baseline methods.
\textbf{Bayesian Additive Regression Trees (BART) \cite{hill2011bayesian}} is a classic tree-based ensemble model.
\textbf{Counterfactual Regression (CFR) \cite{shalit2017estimating}} utilizes two-head neural networks to model potential outcomes, and employs the IPM to address the imbalance issue. 
\textbf{Treatment-agnostic Representation Networks (TARNet) \cite{shalit2017estimating}} is a variant of CFR which does not include the balance regularization.
\textbf{Causal Effect Variational Autoencoder (CEVAE) \cite{louizos2017causal}} is a deep latent-variable model that estimates ITE by modeling the joint distribution. 
 \textbf{Network Deconfounder (ND) \cite{guo2020learning}} uses GCN to capture hidden confounders implicit in networks.
\textbf{Graph Infomax Adversarial Learning (GIAL) \cite{chu2021graph}} improves upon the ND approach by maximizing the mutual information of the network structure.
\textbf {Networked Causal Effects Estimation (NetEst) \cite{jiang2022estimating}} uses GCN to learn confounder representation, and employs adversarial learning to bridge distribution gaps.

\vspace{-0.5em}
\subsection{Experimental Settings}
For all the datasets, we randomly sample 60\% and 20\% of the instances as the training and validation sets, respectively, and the remaining instances are used as the test set. We repeat the simulation procedures 10 times for each dataset with a different imbalance $k$ and report the average results.
For our proposed method, we adopt the RiemannianAdam \cite{chami2019hyperbolic} to optimize the final objective function. Grid search is applied to find the optimal combination of hyperparameters. In particular, we search the learning rate in $\{10^{-3}, 10^{-2} \}$, the absolute value of curvature $c$ in $\{10^{-3}, 10^{-2}, 10^{-1}, 10^{0} \}$, the number of HGCN layers and the number of output regression layers in $\{1,2\}$, 
the dimensionality of HGCN layers and the number of hidden units of  MLP layers in $\{50,100\}$, $\alpha$ in $\{10^{-3}, 10^{-2}, 10^{-1}, 10^{0} \}$, $\beta$ and $\lambda$ in 
$\{10^{-5}, 10^{-4}, 10^{-3}\}$.

To evaluate the performance of the methods, we adopt two widely used metrics, the Rooted Precision in Estimation of Heterogeneous Effect ($\epsilon _{PEHE}$) and Mean Absolute Error on ATE ($\epsilon _{ATE}$), 
where
    $\sqrt{\epsilon_{PEHE}} = \sqrt{\frac{1}{n}\sum_{i=1}^{n}(\widehat{ITE}_i - ITE_i)^2}$
and
   $  \epsilon_{ATE} = |\frac{1}{n}\sum_{i=1}^{n}\widehat{ITE}_i-\frac{1}{n}\sum_{i=1}^{n}{ITE}_i|$. Lower values of these two metrics indicate better model performance.
   
\vspace{-0.5em}
\subsection{Comparison Results}
To demonstrate the effectiveness of our method, we compare the performance of our proposed model with the aforementioned baseline models. We conduct the experiments on the BlogCatalog and Flickr datasets with different settings of $k\in \{1,2,4\}$ as previously described. The results are presented in Table \ref{comparison}, revealing the following observations:
\begin{itemize}[leftmargin=*,topsep=0pt,parsep=0pt,itemsep=0pt]
\item[-]Our method exhibits superior performance compared to other models on both datasets across different settings of $k$, as evident from the lower values of $\sqrt{\epsilon_{PEHE}}$ and $\epsilon_{ATE}$.
Additionally, we conduct a pair-t test, which reveals that our method achieves significantly better performance than the other methods, with a significance level of 0.05.
\item[-]The methods that incorporate network information (ND, GIAL, NetEst, TAHyper) demonstrate superior performance compared to traditional methods. This is because network-enhanced methods account for the presence of hidden confounders and effectively capture them by leveraging the network structure, thereby reducing the confounding bias in ITE estimation. 
\end{itemize}
\vspace{-0.5em}
\subsection{Ablation Study}
To assess the effectiveness of each module in our model, we perform ablation experiments on both datasets, and the corresponding results are presented in Table \ref{ablation}.
Specifically, the ``w/o HB'' denotes the model version where the hyperbolic space is substituted with the Euclidean space. This means that Euclidean features and corresponding Euclidean GCN layers are employed.
Conversely, the ``w/o TA'' signifies the version of the model where the treatment-aware relationship identification module is excluded, and the corresponding term $\alpha \mathcal{L}_t$ in the loss function is also eliminated.
The ablation results indicate that removing either of these two modules leads to a decline in the model's performance, as evidenced by an increase in $\sqrt{\epsilon_{PEHE}}$ and ${\epsilon_{ATE}}$. Therefore, this ablation study conclusively demonstrates that both modules are essential components of our model.
\begin{figure}
\setlength{\abovecaptionskip}{-7pt} 
\setlength{\belowcaptionskip}{-2.0em} 
	\centering
	\begin{minipage}[c]{0.45\textwidth}
		\subfigure{
			\includegraphics[height=3.0cm,width=0.45\columnwidth]{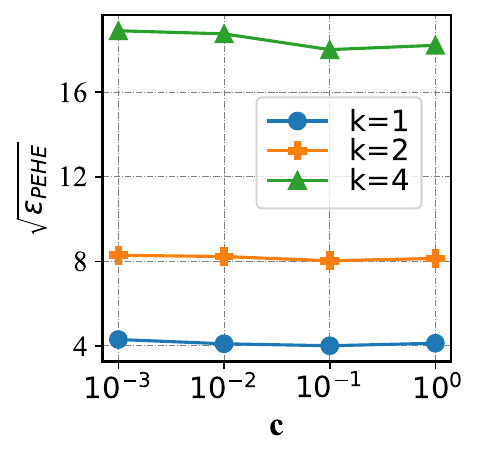} 
			\label{fig11}
   }
   		\subfigure{
			\includegraphics[height=3.0cm,width=0.45\columnwidth]{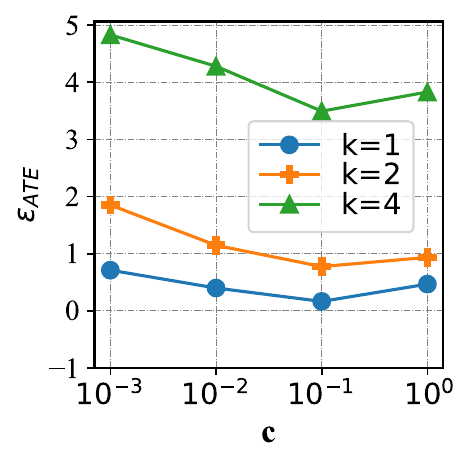} 
			\label{fig12}
	}
  \vspace{-10pt}
	\end{minipage}

	\begin{minipage}[c]{0.45\textwidth}
		\subfigure{
			\includegraphics[height=3cm, width=0.45\columnwidth]{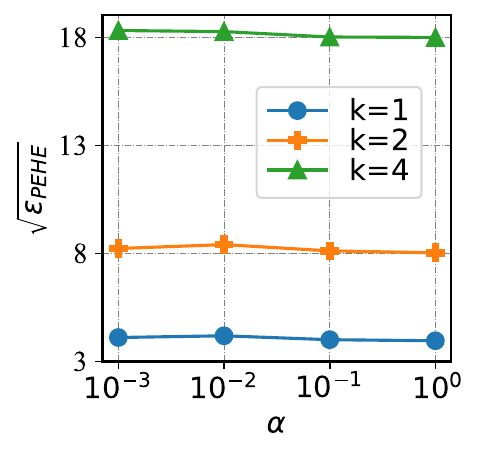} 
			\label{fig13}
   }
   		\subfigure{
			\includegraphics[height=3cm, width=0.45\columnwidth]{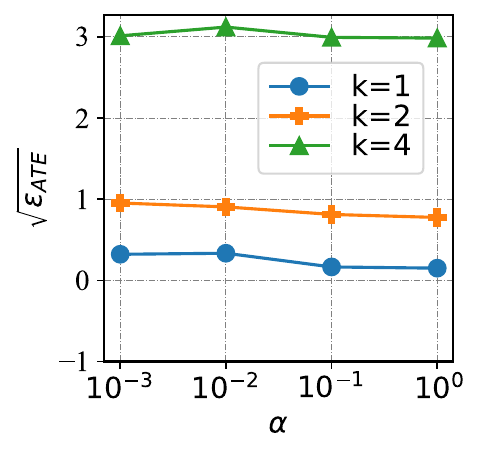} 
			\label{fig14}
	}
	\end{minipage}
	\caption{Hyperparameter study on BlogCatalog.}
	\label{hyps}
\end{figure}

\vspace{-0.5em}
\subsection{Hyperparameter Study}
We investigate the impact of two important hyperparameters, namely $c$ (the absolute value of the negative curvature) and $\alpha$ (the coefficient in the loss function), on the performance of our model.
The results are shown in Fig. \ref{hyps}. We only present the results on the BlogCatalog dataset due to space limitations. 
As $c$ and $\alpha$ vary, the values of $\sqrt{\epsilon_{PEHE}}$ and ${\epsilon_{ATE}}$ exhibit minimal changes, suggesting the insensitivity of our model to these hyperparameters.
In particular, $c$ reflects how much a geometric object deviates from a flat plane, where $c=0$ signifies an Euclidean space. 
From our experimental results, increasing $c$ within the range of $10^{-3}$ to $10^{-1}$ leads to a gradual decrease in two metrics, indicating an improvement in model performance. 
However, when $c$ surpasses a certain threshold (greater than 0.1), a slight decline in performance is observed. This implies determining the suitable curvature for a given dataset can unleash the maximum capabilities of the hyperbolic space.

\vspace{-0.5em}
\section{Conclusion}
This paper presents a novel method called treatment-aware hyperbolic representation learning for ITE estimation. Our model integrates two innovative designs, aiming to effectively capture hidden confounders with social networks. Firstly, we employ hyperbolic space to encode the network structure, reducing the distortion caused by Euclidean space and improving the quality of confounder representation from a macro perspective. Secondly, we enhance the hidden confounder learning from an individual view by incorporating a treatment-aware relationship identification module. The superiority of our method is demonstrated through extensive experiments on two benchmark datasets.
\vspace{-0.5em}
\section*{Acknowledgement}
This work was supported by the Start-up Grant (No. 9610564) and the Strategic Research Grant (No. 7005847) of the City University of Hong Kong.
\vspace{-0.5em}
\begin{spacing}{0.9} 
\bibliographystyle{siam}
\bibliography{reference}
\end{spacing}

\end{document}